\documentclass{article}

\PassOptionsToPackage{numbers, compress}{natbib}
\usepackage[preprint]{neurips_data_2024}

\usepackage[utf8]{inputenc} % allow utf-8 input
\usepackage[T1]{fontenc}    % use 8-bit T1 fonts
\usepackage{hyperref}       % hyperlinks
\usepackage{url}            % simple URL typesetting
\usepackage{booktabs}       % professional-quality tables
\usepackage{amsfonts}       % blackboard math symbols
\usepackage{nicefrac}       % compact symbols for 1/2, etc.
\usepackage{microtype}      % microtypography
\usepackage{xcolor}         % colors

\usepackage[pdftex]{graphicx}
\usepackage{CJKutf8}
\usepackage{enumitem}
\usepackage{multicol}
\usepackage{multirow}
\usepackage{xspace}
\usepackage{amssymb}
\usepackage{epigraph}
\usepackage{mathtools}
\usepackage{changepage}
\usepackage{listings}
\title{Creating a Lens of Chinese Culture: A Multimodal Dataset for Chinese Pun Rebus Art Understanding}

\author{%
  Tuo Zhang$^{1}$\thanks{These authors contribute equally}\footnotemark[1] , Tiantian Feng$^{1}$\footnotemark[1], Yibin Ni$^{2}$\footnotemark[1], Mengqin Cao$^{3}$\footnotemark[1], \\
  \textbf{Ruying Liu$^{1}$, Katharine Butler$^{4}$, Yanjun Weng$^{6}$} \\
  \textbf{Mi Zhang$^{5}$, Shrikanth S. Narayanan$^{1}$, Salman Avestimehr$^{1}$}
  \\
$^1$University of Southern California, $^2$Shanghai International Studies University, \\
$^3$Independent Researcher, $^4$The Butler Museum, $^5$The Ohio State University\\
$^6$Jingdezhen Imperial Kiln Institute}

\begin{document}

\maketitle

\begin{abstract}
Large vision-language models (VLMs) have demonstrated remarkable abilities in understanding everyday content. However, their performance in the domain of art, particularly culturally rich art forms, remains less explored. As a pearl of human wisdom and creativity, art encapsulates complex cultural narratives and symbolism. In this paper, we offer the Pun Rebus Art Dataset, a multimodal dataset for art understanding deeply rooted in traditional Chinese culture. We focus on three primary tasks: identifying salient visual elements, matching elements with their symbolic meanings, and explanations for the conveyed messages. Our evaluation reveals that state-of-the-art VLMs struggle with these tasks, often providing biased and hallucinated explanations and showing limited improvement through in-context learning. By releasing the Pun Rebus Art Dataset, we aim to facilitate the development of VLMs that can better understand and interpret culturally specific content, promoting greater inclusiveness beyond English-based corpora.
\end{abstract}

\begin{CJK*}{UTF8}{gbsn}

\section{Introduction}
\label{sec:introduction}

\begin{figure}[t]
 \vspace{-7mm}
 \centering
 \includegraphics[width = \linewidth]{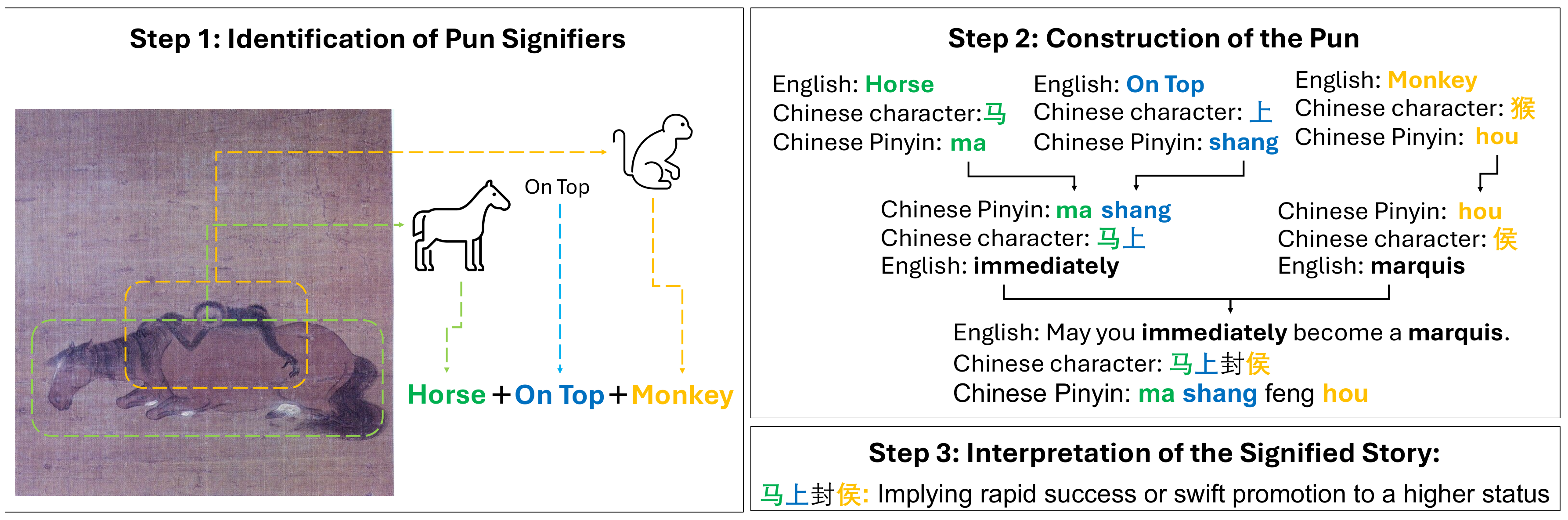}
    \caption{The illustration of the chain of thought on understanding the Chinese pun rebus. The example artwork uses a horse and a monkey to construct the pun "马上封侯" (mǎ shàng fēng hóu), which means "May you instantly become a marquis" in English.}
    \label{fig:cot_data}
    \vspace{-4mm}
\end{figure}

Each culture develops its unique symbolic systems of visual elements, which are conventionally understood within that culture to convey specific meanings. For example, to viewers unfamiliar with Chinese arts and linguistics, the combination of a monkey and a horse might seem nonsensical. However, in Chinese culture, "a monkey lying on top of the horse" is described as a pun on "马上封侯" (mǎ shàng fēng hóu)\footnote{The notation is in Pinyin, the official romanization pronunciation system for Standard Mandarin Chinese}, representing the wish for promotion. This form of wordplay is prevalent in Chinese decorative arts, appearing in various art formats throughout Chinese history, from the emperor’s court to the commoners’ kitchen, transcending boundaries of power, wealth, education, and media. As an example, in Figure~\ref{fig:cot_data}, we demonstrate a Chinese pun rebus painting with "a monkey lying on top of the horse," which indicates the wish for promotion by connecting homophonically similar Chinese characters of "horse-马(mǎ)," "on top of-上(shàng)," combined to form ‘mashang’ also meaning ‘right away’, and "monkey-猴(hóu), sounding similar to 侯(hóu) for ‘marquis." 
% of "horse-马 (mǎ)," "on top of-上 (shàng)," and "monkey-猴 (hóu)$\rightarrow$侯 (hóu)." 

In this work, we propose the Pun Rebus Art Dataset, which is rooted in traditional Chinese culture. We focus on Chinese Pun Rebus art for three major reasons: 1) creating a pun rebus artwork involves combining textual meanings with corresponding visual representations, making it naturally multimodal; 2) pun rebus is prevalent in Chinese art, rarely seen in other cultures such as western painting~\cite{yibin2003anatomy}; 3) pun rebus art remains widespread in contemporary Chinese culture, demonstrating its enduring impact and lasting value to preserve cultural identity while engaging new generations. 

We introduce three sequential tasks that reflect the underlying \textit{chain-of-thought} process of experts in decoding Chinese pun rebuses. Our goal is to benchmark the capability of large vision-language models (VLMs) in recognizing, interpreting, and comprehending these rich and cultural-specific meanings across vision and language: 1) identifying the salient and relevant visual elements in art; 2) matching the visual elements with symbolic meaning; and 3) generating an explanation to express why an artwork convey certain messages. To the best of our knowledge, this is one of the first datasets to test AI's abilities in handling cultural-specific art expression, particularly focusing on the accurate identification and interpretation of visual signifiers within Chinese pun rebus art.

Our results highlight the inherent challenges by both AI and non-expert humans in understanding Chinese pun rebus arts compared to experts. In the visual element identification task, even the best VLM captures only about 30\% of key elements, slightly outperforming non-expert humans. Moreover, most VLMs struggle to match the symbolic meanings associated with Chinese culture, with GPT-4o achieving the highest accuracy of 42\% in a 7-way multiple-choice question. In comparison, non-expert humans manage to reach a 55\% accuracy in this task. Finally, experts note that the explanations generated by VLMs in expression understanding often involve biases and hallucinations, underscoring the current VLMs' limitation in understanding Chinese art and potentially other culturally specific contents. We hope that our effort in curating, releasing, and benchmarking the Chinese Pun Rebus Arts dataset will facilitate the development of VLMs in understanding cross-cultural content other than English-based corpus, thereby promoting greater inclusiveness.

\end{CJK*}

\begin{CJK*}{UTF8}{gbsn}

\vspace{-1mm}
\section{General Framework for Pun Rebus Understanding} \label{sec:pun}
\vspace{-1mm}

% \subsection{General Framework for Pun Rebus Understanding}
A pun rebus in Chinese culture leverages visual elements to indicate an underlying expression, metaphor, or meaning that is seemingly unrelated to the given image~\cite{yibin2003anatomy, yibinbbook}. The fundamental mechanism of pun rebuses hinges on the interplay between the imagery composed, on the one hand, and the semantic and phonetic components of the Chinese logographs used to express a message, usually auspicious. Specifically, the interpretation of pun rebuses relies on homophonic associations between the names of the depicted images (or their interactions) and the Chinese characters (logographs) used to express the concepts that form the intended message, either partially or fully. The names of the objects in a pun rebus are often homophonically similar to, or even identical with, the cued expression, analogous to using the English string ‘eye—can—sea—ewe’ to express ‘I can see you’. A pun rebus design is intended to initiate a cognitive translation process of "image-sound-sound-meaning," contrasting sharply with the more direct and straightforward ‘text-meaning’ decoding typically observed in pure verbal understanding. Because the process is not only culturally but also linguistically specific, it is extremely challenging for an uninformed viewer to perceive and decipher any underlying meanings of this art form. These artworks are composed for aesthetic or attention-attracting purposes.

Generally, the chain of thought on understanding the pun rebus is composed of three sequential steps:
(1) spotting the salient visual elements within the artwork; (2) utilizing these identified elements to formulate the underlying pun; (3) understanding the intended message or wish conveyed by the pun rebus.
We present a visualized example in Figure~\ref{fig:cot_data} as an illustration of pun rebus understanding.

\end{CJK*}
% A framework of the general protocols for the designing of pun rebuses is presented below.

% The framework consists of five aspects to account for different methods involved in expressing meaning through punning rebuses:

% [1] relationship between images used and concepts conveyed
% [2] modes of relationship between images
% [3] ways of punning: different levels of similarity between the sound of the signifier and that of the signified
% [4] degrees of explicitness in verbal explanation
% [5] ways of deriving the pun signifier
% [6] relationship between the pun signified and the intended message

% [1.0] The relationship between images and concepts

% On the one hand, the same image may mean different things in different visual contexts or in the company of different objects; the image of an object may serve as a cue for more than one concept. On the other hand, different images may cue the same concept in different visual contexts.

\vspace{-1mm}
\section{Pun Rebus Art Dataset} \label{sec:dataset}
\vspace{-1mm}
\vspace{-1mm}
\subsection{Data Collection}
\vspace{-1mm}
The Pun Rebus Art dataset is designed as a comprehensive benchmark for exploring the intersection of image analysis, morphological variation, and phonological elements within the context of Chinese linguistics and cultural artifacts. This dataset is the result of extensive efforts to curate a diverse array of historical artwork documents. Initiated in 1987 by Dr. Ni Yibin, a co-author of this paper, the dataset’s preparation involved meticulous collection, annotation, and verification processes that require expert knowledge of Chinese art, literature, history, and linguistics. The corpus comprises 1,011 captioned images sourced predominantly from globally-renowned Chinese-art-collecting institutions, including the Palace Museum, the Metropolitan Museum of Art, and the British Museum. The images in this dataset are subject to the Creative Commons Zero (CC0) license. Spanning over two millennia, from the Han Dynasty (206 BCE – 220 CE) to the 20th century, the dataset encompasses a rich diversity of more than ten different media types, including paintings, ceramics, bronzes, sculptures, jade, Cloisonné, lacquerware, and embroidery. The collection of the Pun Rebus Art dataset is ongoing as we continue to curate it with additional artworks to enhance its representational diversity. 

\subsection{Data Annotation} \label{sec:data annot}
\begin{figure}[t]
\vspace{-10mm}
 \centering
 \includegraphics[width = \linewidth]
 {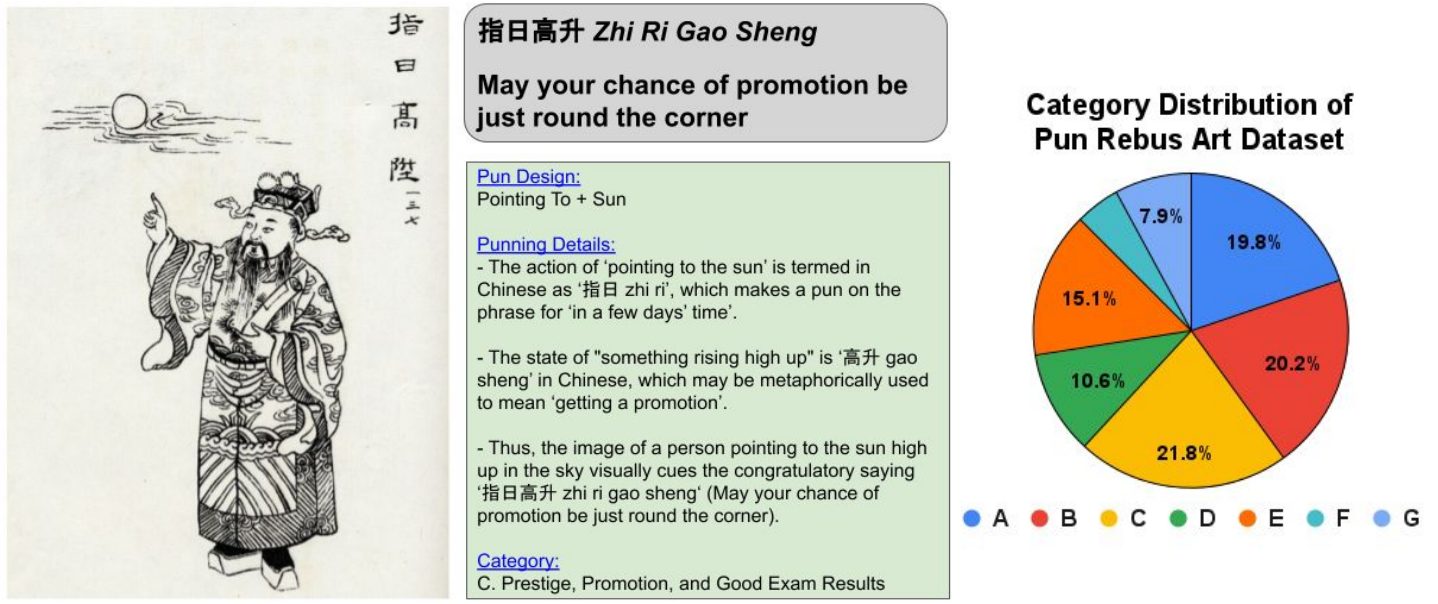}
 % {figures/example_sample.pdf}
    % \vspace{-4mm}
\caption{An example data sample and category distribution of the Pun Rebus Art dataset. We offer both English and Chinese versions of the data annotation in the proposed dataset. The dataset querying system is available on \url{http://niyibin.org/punrebus/punrebus_main_en.php}.}
\label{fig:data_example}
\vspace{-4mm}
\end{figure}

Each entry has been meticulously annotated by human experts with knowledge of Chinese linguistics, art, and history. Figure~\ref{fig:data_example} exemplifies the structured content in the Pun Rebus Art dataset. Each entry comprises the following components: (1) the original artwork without its caption; (2) the articulated pun rebus, presented bilingually to encompass both the original Chinese script and its English counterpart; (3) the salient elements that constitute the pun's design; and (4) an analysis delineating the relationship between the visual representation and the intended pun rebus. To ensure high-quality annotations, we implement a strict three-round validation check after the initial annotating process.

\vspace{-1mm}
\section{Task Setups} \label{sec:task}
\vspace{-1mm}
Based on the characteristics of the Chinese pun rebus artwork, we present three primary and progressive tasks in this paper: \textit{Element Identification}, \textit{Symbolic Matching}, and \textit{Expression Understanding}. We want to highlight that the researchers are highly encouraged to explore additional applications and analyses tailored to their specific interests and needs using this dataset. In the following, we describe the details of each task and the corresponding evaluation metrics.
% \vspace{-1mm}
\subsection{Task Design}
\begin{figure*}[t]
 \vspace{-10mm}
 \centering
 \includegraphics[width = \linewidth]{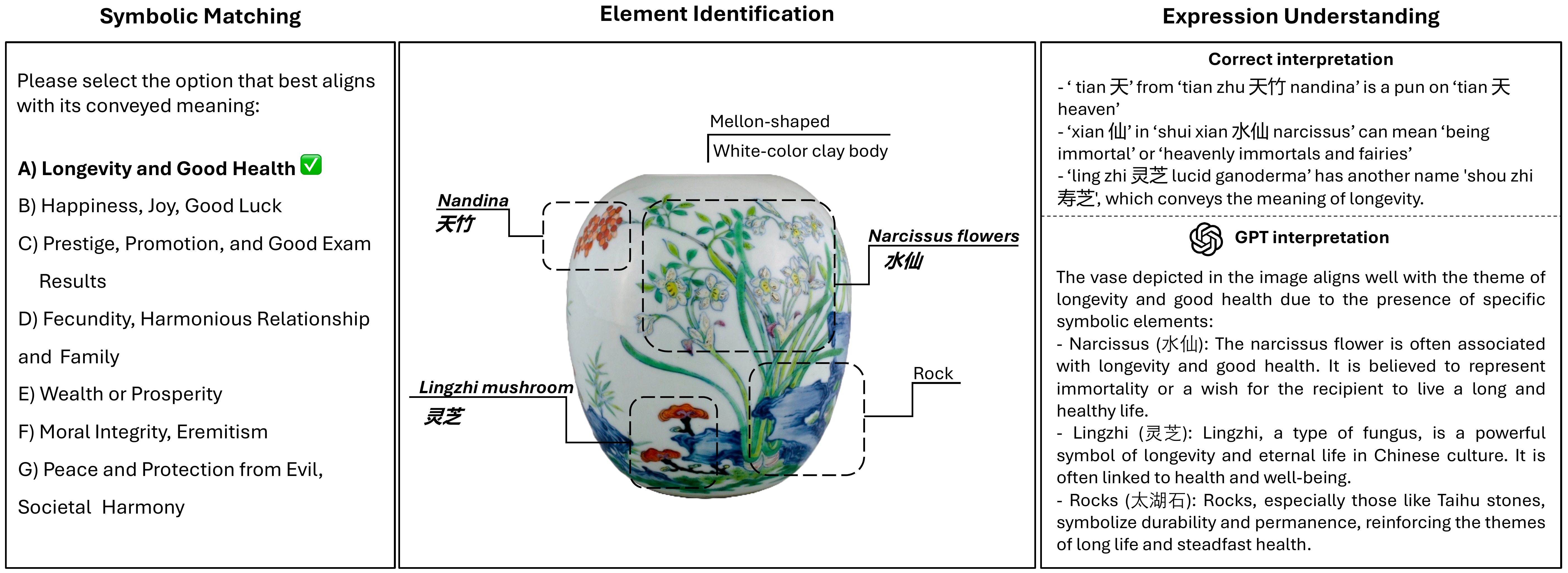}
    \caption{Illustration of the three evaluation tasks using an 18th-century Chinese ceramic as an example. \textbf{Bold} marks the salient elements. \textit{Symbolic Matching} probes the model's understanding of the artwork's implied meanings. \textit{Element Identification} asks what catches the model's attention most in the artwork. \textit{Expression Understanding} delves into the rationale behind the model's interpretations.}
    \label{fig:task}
    \vspace{-4mm}
\end{figure*}

\vspace{-1mm}
\textbf{Element Identification.}
In the initial task, we aim to explore: \textit{What catches the model's attention most in the artwork?}
Artworks are complex composites of features such as texture, shape, color, and other painting elements. However, not all these features are essential for constructing the pun embedded within the artwork. This task seeks to determine which elements the model prioritizes from its perspective. For instance, consider the artwork shown in Figure~\ref{fig:task}: a ceramic jar made in the Yongzheng period of the Qing Dynasty (1732 - 1735). This jar exhibits numerous features, including its egg-like shape, the white-color clay body, the flowers at the top, and the colorful rock at the bottom. However, only the narcissus flowers, the red berries of nandina, and the lingzhi mushrooms depicted on the jar are crucial to its implied wishes. In Chinese, the sound of 'narcissus' echoes the phrase for 'heavenly immortals and fairies,' and the sound of 'nandina' echoes 'heaven,' while 'lingzhi mushrooms' are traditionally associated with longevity. Their combined presence suggests the wish 'May you enjoy a long life as immortals.' In contrast, other elements like the jar's shape or the rock at the bottom, while visually striking, do not contribute significantly to the articulated wish.

\textbf{Symbolic Matching.} 
In the second task, we investigate the question: \textit{What does the model understand after reading the artwork?} 
Drawing upon expertise in Chinese iconographic art history and cultural studies, we have classified the auspicious expressions depicted in the datasets into seven categories, as the example shown in Figure~\ref{fig:task}. The category distribution is presented in Figure~\ref{fig:data_example}. We ask the model to make a selection that best aligns with the conveyed meaning behind the given artwork images among the eight options. This task serves as a direct evaluation of the model's ability to comprehend the pun rebus reasoning embedded within each artwork.

\textbf{Expression Understanding.}
Finally, we want to see \textit{Why does the model interpret the artwork as it does?}
This task is designed to delve into the reasoning behind the model’s decisions, providing insights into its interpretative process. By understanding the justifications for the model’s choices, we can assess how closely it aligns with human understanding of cultural and symbolic meanings. 
\vspace{-2mm}
\subsection{Evaluation Metrics} \label{sec:eva_metrics}
% \textbf{Symbolic Matching.} For the symbolic matching, we evaluate using the accuracy. It is worth noting that certain artworks may convey multiple implied meanings among the options provided. An answer is considered correct if it includes at least one implied meaning specified in the ground truth.

\textbf{Element Identification.} For the element identification, we report the absolute score and the similarity score. The absolute score represents the overlap between key elements in the model’s output and those in the ground truth. Let $G = \{g_1, g_2, \dots, g_n\}$ represent the set of elements in the ground truth description, and $P = \{p_1, p_2, \dots, p_m\}$ denote the set of elements identified by the language model. The absolute score for a single instance, $S$, is calculated as follows:
\begin{equation}
    S_{Abs}(G, P) = \frac{|G \cap P|}{|G|}
\end{equation}
It quantifies the extent to which essential elements are captured in the model's output, normalized by the total number of elements in the ground truth. For the overall performance across the dataset, we report the average score, $\overline{S_{Abs}}$, computed as the mean of individual scores across all test instances. 

Apart from the absolute score, we introduce the similarity score to account for semantic equivalence, which considers synonyms and semantically related terms that align with the ground truth. We map both ground truth answers $G$ and generated answers $P$ to word embedding using the pre-trained Sentence-BERT~\cite{reimers-2019-sentence-bert}. For each test instance, we measure word-wise cosine similarity between each element in $G$ and all elements in $P$, recording the highest similarity score for each element in $G$. The similarity score for each instance is the average of these maximum scores for all elements in $G$:
\begin{equation}
    S_{Sim}(G, P) = \frac{1}{|G|} \sum_{g \in G} \max_{p \in P} \cos(emb(g), emb(p))
\end{equation}
where $emb(x)$ denotes the embedding of the element $x$, and $\cos$ denotes the cosine similarity function. We report the average score $\overline{S_{Sim}}$ for the overall performance of the dataset.

% This metric captures semantic correspondences between the model’s outputs and the ground truth.

\textbf{Symbolic Matching.} For the symbolic matching, we evaluate using the accuracy. It is worth noting that certain artworks may convey multiple implied meanings among the options provided. An answer is considered correct if it includes at least one implied meaning specified in the ground truth.

\textbf{Expression Understanding.} We conduct human evaluation to judge the expression understanding. The panel of human judges consists of five individuals: three authors of this paper and two independent experts with educational and professional backgrounds in the field of art history. We ask each judge to grade the model-generated explanations on a scale from 1 to 10. A score of 10 represents a perfect explanation, indicating that the human judge cannot distinguish whether the answer is from the machine or a human expert. A score of 1 signifies that the response is completely incorrect and irrelevant. We also list our findings and hypothesis from human evaluations in Section~\ref{sec:human_check}.

\section{Experiments}
We evaluate the performance of various widely used VLMs using the Pun Rebus Art dataset. 
Our evaluation is conducted under both zero-shot and five-shot settings to examine the inherent ability without fine-tuning specific to our dataset.
Specifically, we aim to probe the ingrained knowledge and reasoning processes of these models, exploring their potential limitations or biases in interpreting objects and concepts related to Chinese culture. This is particularly pertinent to ensure the inclusiveness of VLMs given that most models are predominantly trained on English-based resources, which may affect their performance on culturally specific tasks~\cite{zhang-etal-2023-dont, huang2023not}. 
We use the unified prompt for each task across all models, which are listed in Appendix. We sample with default hyperparameters in all cases. All experiments are conducted with NVIDIA A100 GPUs.

\subsection{Baselines}
\textbf{VLMs.} 
Our selection prioritizes the largest, most recent, and highest-performing VLMs currently available. Our selection comprises:
(1) The GPT-4 model family~\cite{achiam2023gpt}. We include both GPT-4o and GPT-4V in our benchmark. 
(2) The Gemini 1.0 Pro Vision~\cite{team2023gemini} is the only multimodality LMM available for public usage among the Gemini model family.
(3) The Claude 3 model family~\cite{anthropic2024claude}. Our benchmark includes Claude 3 Opus, Sonnet, and Haiku.
(4) The Qwen-VL family~\cite{bai2023qwen}. It is worth noting that its training incorporates the Chinese image-text data corpus, making it more relevant to benchmark our dataset. We incorporate both Qwen-VL-Plus and Qwen-VL-Max in our evaluation.
For all models listed, we utilize the latest model checkpoint available at the time of writing this paper. Detailed checkpoint information and version specifics are provided in Appendix.

\textbf{Human Performance Estimates.} 
Following the previous study~\cite{hessel2023androids}, we include an evaluation of human performance to compare with the VLMs. Unlike the expert panel described in Section~\ref{sec:eva_metrics}, we enlist crowd-workers who lack a specialized background in Chinese art, representing the general population's understanding. Specifically, the panel consists of 3 bilingual individuals, all native Chinese speakers who are also fluent in English. Each participant will review 50 artworks and respond to the questions related to symbolic matching and element identification. These artworks are randomly selected in the full dataset with the same label distribution. We report the average scores across participants as the human performance estimates. We want to note that the human performance in this paper should not be considered as an upper bound for VLMs. Instead, it is used to measure how well ordinary people raised in contemporary Chinese society understand traditional Chinese arts.

\newcommand{\zshotres}{\hspace*{.4in}\rotatebox[origin=c]{180}{$\Lsh$}\xspace}
\newcommand{\fiveshotres}{\zshotres}
\begin{table*}[t]
\vspace{-10mm}
\begin{minipage}[t]{0.6\textwidth}
\resizebox{\textwidth}{!}{
\begin{tabular}{lccc}
& \multicolumn{1}{c}{Symbolic Matching} & \multicolumn{2}{c}{Element Identification} \\
\cmidrule(lr){2-2}\cmidrule(lr){3-4}
& Accuracy ($\uparrow$) & $\overline{S_{Abs}}$ ($\uparrow$) & $\overline{S_{Sim}}$ ($\uparrow$)\\
\midrule
Random Choice & 14.29\% & - - & - - \\
\midrule
GPT-4o & \textbf{40.40\%} & 0.3145 & \textbf{0.5688}  \\
\quad \rotatebox[origin=c]{180}{$\Lsh$} Five-shot & \zshotres 42.18\% & \zshotres 0.3499 & \zshotres 0.5851 \\
GPT-4V & 26.53\% & 0.2616 & 0.5003 \\
Gemini Pro & 27.92\% & \textbf{0.3398} & 0.5003 \\
Claude 3 Opus & 22.47\% & 0.2405 & 0.4983 \\
\quad \rotatebox[origin=c]{180}{$\Lsh$} Five-shot & \zshotres 19.77\% & \zshotres 0.2623 & \zshotres 0.5127 \\
Claude 3 Sonnet & 20.55\% & 0.1767 & 0.4030 \\
Claude 3 Haiku & 21.91\% & 0.1713 & 0.4350 \\
Qwen-VL-Max & 37.77\% & 0.2453 & 0.4786 \\
\quad \rotatebox[origin=c]{180}{$\Lsh$} Five-shot & \zshotres 21.45\% & \zshotres 0.0327 & \zshotres 0.3406 \\
% 22.99\%
Qwen-VL-Plus & 28.88\% & 0.2545 & 0.4131 \\
\midrule
Human Estimate & 55.33\% & 0.2483 & 0.4615 \\
\bottomrule
\end{tabular}
}
\end{minipage}
\begin{minipage}{0.4\textwidth}
\centering
\includegraphics[width=.9\linewidth]{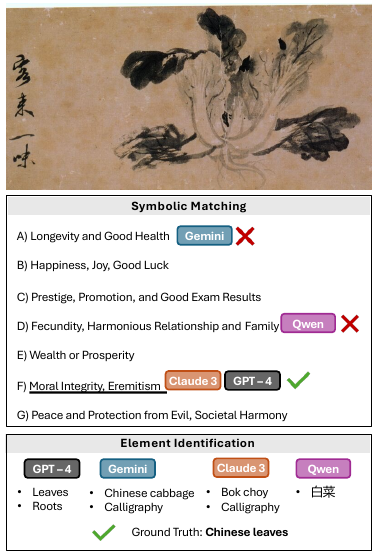}
\end{minipage}
\caption{Evaluation results for the symbolic matching and element identification tasks among various VLMs. \textbf{Bold} results are best for zero-shot evaluation in each category.
Right: sample results by GPT-4o, Gemini Pro, Claude 3 Opus and Qwen-VL-Max over a matching/identification instance. 
} 
\label{tab:main_results}
\vspace{-4mm}
\end{table*}

\vspace{-2mm}
\subsection{Main Results} \label{sec:main_results}
\vspace{-1.5mm}
\subsubsection{Evaluation under Zero-shot Settings} 
\vspace{-1.5mm}
In this section, we compare different VLMs through a zero-shot evaluation of the Pun Rebus Art benchmark, as detailed in Table~\ref{tab:main_results}. We make five key observations:

(1) \textit{The challenging nature of the Pun Rebus Art dataset.} 
We observe that the highest accuracy in symbolic matching achieved under the zero-shot setting is around 40\% for all models. Notably, the human estimation also averages only around 55\%, underscoring the difficulty of understanding the symbolic meaning in the art. As we stated in Section~\ref{sec:dataset}, the Pun Rebus Art dataset spans artwork ranging over 2000 years, where many visual representations or underlying narratives may have lost their prominence in contemporary Chinese culture. Moreover, to correctly understand an artwork, VLMs must first identify the key elements and then connect these elements into a coherent story. 

% The vast knowledge required and the intricate chain of thought for reasoning make the Pun Rebus Art dataset a unique resource for evaluating the capabilities of VLMs in cross-cultural understanding.

% This cognitive process is akin to how humans understand and interpret art, where recognizing individual components such as figures, symbols, and stylistic nuances forms the foundation of subsequent reasoning.

(2) \textit{The Pun Rebus dataset extends beyond the knowledge scope of VLMs.}
The relatively low scores observed in the element identification reveal that the tested VLMs fail to understand the Pun Rebus artworks, leading to 50\% of the key elements being missed in the recognition. The even lower accuracy in the symbolic matching reflects VLMs' sparse knowledge of Pun Rebus-related content, demonstrating their lack of sufficient knowledge and reasoning ability to transfer the identified key elements into the conveyed meanings. The substantial historical span of the dataset, combined with the struggling performance observed in our evaluations, indicates that the cultural and linguistic content within these artworks extends beyond the training knowledge of the tested models. 

(3) \textit{Element recognition versus cultural interpretation limits VLMs.}  
The VLM with high symbolic matching accuracy, such as GPT-4o, also scores well in element identification. However, VLMs like Claude 3 Opus score high in element recognition but struggle with symbolic understanding. For example, as we showed in Table~\ref{tab:main_results}, Claude 3 Opus identifies bok choy in the artwork but fails to link it to moral integrity, a symbol in Chinese culture due to its similar pronunciation with 'incorruptible.' This highlights a critical aspect of VLM's performance: translating visual recognition into meaningful cultural interpretation. In Appendix, we detail 12 distinct mechanisms used in Chinese culture to derive symbolic meanings from visual elements, including puns, shapes, numerals, and aliases. 
% Thus, the Pun Rebus Art dataset evaluates VLMs not only on recognition but also on reasoning, serving as a benchmark for multimodal understanding in culturally rich contexts.

(4) \textit{GPT-4o demonstrates superior performance compared to other models.} Notably, GPT-4o largely outperforms other models in our evaluation, including GPT-4V. This improvement is partly due to enhanced visual recognition abilities, as evidenced by higher element identification scores achieved by GPT-4o compared to GPT-4v. Other factors, such as the integration of end-to-end multimodal learning techniques in GPT-4o, may also lead to a more effective interpretation of complex visual and textual information. Despite these notable improvements, the precise factors contributing to the improved performance of GPT-4o remain unclear to us.

% This could contribute to its improved performance in understanding and reasoning about the cultural symbols in the Pun Rebus Art dataset.

% The precise factors contributing to the improved performance of GPT-4o remain unclear to us. 

% Additionally, the integration of advanced multimodal learning techniques in GPT-4o allows for a more effective combination and interpretation of visual and textual information, contributing to its superior performance in understanding and reasoning about the cultural symbols in the Pun Rebus dataset.

(5) \textit{The impact of using Chinese image-text data corpus in the Pre-training of VLMs.}
Among the tested models, only the Qwen-VL family publicly announced substantial Chinese data in their training corpus. Our Pun Rebus dataset is naturally bilingual, with content rooted in Chinese culture and questions posed in English. 
The Qwen-VL-Max achieved the second-highest accuracy in symbolic matching, only below GPT-4o. Examination of Qwen model responses in element identification showed that 18.99\% of Qwen-VL-Max and 17.90\% of Qwen-VL-Pro responses were in Chinese characters. This language mismatch contributed to their relatively low scores in element identification, as the ground truth answers were in English. Appendix includes examples of these responses with corresponding artwork images. Human inspection further found that Chinese responses predominantly occurred with images deeply embedded in Chinese culture, such as traditional ink paintings or fable stories. We speculate that the Qwen models were exposed to Chinese culture-related image-text pairs without English translations during their pre-training. Consequently, the models defaulted to Chinese responses instead of English when encountering similar elements.

\subsubsection{Evaluation under Few-shot Settings}
We also evaluate the in-context learning ability of models using a 5-shot prompt on the Pun Rebus dataset. Specifically, we select the best-performing models from each model family: GPT-4o, Claude 3 Opus, and Qwen-VL-Max. We do not include Gemini Pro because the currently publicly available API only supports interleaved images as a few shot prompts but not the multiple image input as the other model. The results are presented in Table 1. We make two key observations:

(1) \textit{Marginal improvements with five-shot prompting.} With five-shot prompting, we observed slight increases in the symbolic matching performance of GPT-4o and in the element identification performance of both Claude 3 Opus and GPT-4o. 
%
% The improvement in element identification suggests that incorporating demonstrating examples within the prompt helps guide the VLM models in recognizing the visual elements better. 
The prompt directly illustrates what the elements look like and highlights which elements are important to the conveyed meaning, leading to improved performance in element identification.
However, element identification is inherently simpler and requires less reasoning compared to symbolic matching. Symbolic matching is more complex, as the model must identify the mechanisms to integrate the spotted elements into coherent stories. 
The prompts provided answers but did not explain the underlying mechanisms, resulting in minimal improvement in symbolic matching. In some cases, the performance is even lower compared to zero-shot settings, as the model could not understand the reasoning behind the prompts. 

% The substantial outperformance of GPT-4o over Claude 3 Opus highlights their differing abilities in reasoning, with GPT-4o demonstrating superior language reasoning capability, particularly in the context of Chinese culture.

(2) \textit{Hallucination and Shortcuts Exploitation to the in-context examples.} With Qwen-VL-Max, we observe the performance decreases in all tasks under the five-shot settings. With human inspection of the element identification responses, we found that the word "Pheasant" appeared 317 times, approximately 31.35\% of all answers. In our provided prompt, we included an example labeled "Quail." Biologically, quails belong to the pheasant family. We speculate that the behavior of the large VLMs is associated with the "lazy learners" phenomenon discussed in \cite{tang2023large}, where the VLMs frequently
exploit shortcuts in in-context examples for downstream
tasks. This leads the model to incorrectly identify various elements as "Pheasant," regardless of whether the artwork depicted humans, flowers, or other animals. These observations suggest that the VLMs tend to exploit shortcuts to the in-context examples, resulting in the generation of hallucinated answers that are close to the few-shot examples, which could be the major reason for the decline in performance.

% The responses no longer contained Chinese characters and were entirely presented in English. 

% We speculate that the model overfit to this example, leading it to incorrectly identify various elements as "Pheasant," regardless of whether the artwork depicted humans, flowers, or other animals. Additionally, the responses no longer contained Chinese characters and were exclusively in English. These observations suggest that the model may overfit to the given examples and produce hallucinated answers, which is the major reason for the performance declines.

\subsection{Human Evaluation and Error Analysis} \label{sec:human_check}

\begin{figure}[t]
\vspace{-0mm}
 \centering
 \includegraphics[width = \linewidth]{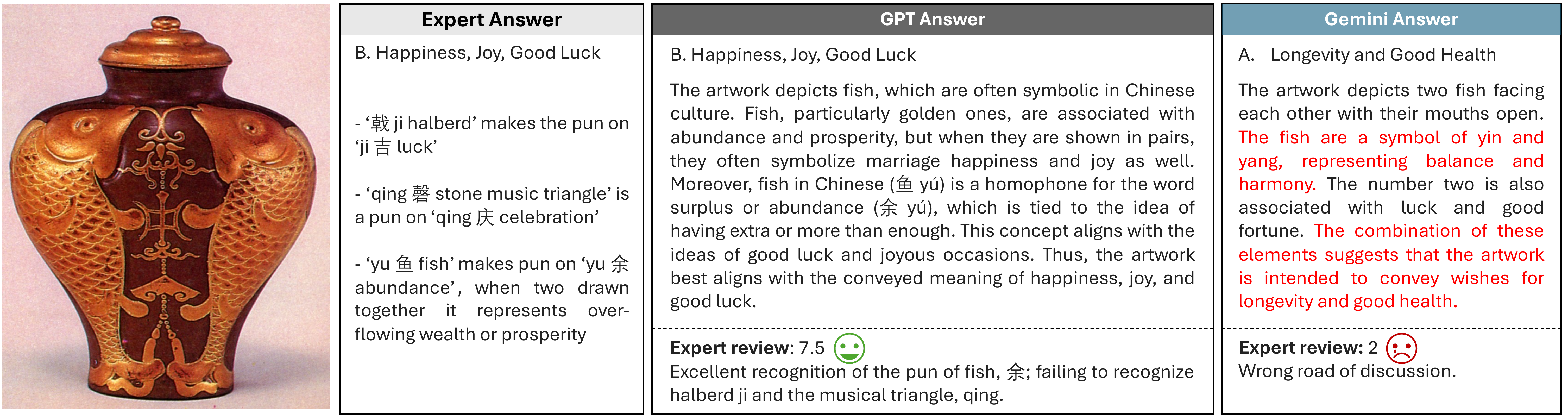}
    % \vspace{-4mm}
\caption{An example of expression understanding generated by GPT-4o and Gemini Pro, including the expert review and the expert-provided answer for this artwork. Errors are highlighted in red.}
\label{fig:expert}
\vspace{-4mm}
\end{figure}

\subsubsection{Expert Review on Expression Understanding}
Our expert judges reviewed the expression understanding generated by GPT-4o and Gemini Pro. We randomly selected 50 responses from each VLM, ensuring the samples maintained the same category distribution as the full dataset. An example review and the judges' explanations are shown in Figure~\ref{fig:expert}. Overall, GPT-4o received an average score of 3.47, while Gemini Pro received an average score of 3.01 from the expert judges. The expert judges make two key observations from the reviews:    

(1) \textit{Reasons of errors in expression understanding.} The primary issue is incorrect recognition or missing salient elements. For example, both models failed to recognize a persimmon in one artwork, mistakenly identifying it as a peach, reflecting the challenges in element identification shown in Table~\ref{tab:main_results}. 
Secondly, even when VLMs correctly identified elements, they often misunderstood the conveyed meaning. As shown in Figure~\ref{fig:expert}, Gemini recognized fish but completely missed its pun. Also, Gemini tends to fabricate things that do not appear in the pun rebus designs. Lastly, in some cases, the VLMs achieved an expert-level understanding but selected an incorrect option.

% Expert reviews highlight several reasons for the errors made by the models. 

(2) \textit{Potential bias in VLMs.} Experts noted potential bias in the generated answers. When VLMs fail to recognize an element, they tend to link it to common symbols in Chinese culture, specifically bats, peaches, pine trees, and rocks, which are frequently used to represent longevity and good luck. They often defaulted to associating uncertain elements with these four elements based on shape similarity. For example, they might interpret long, tree-shaped elements as pine trees and round-shaped elements as peaches based on shape similarity.  Additionally, VLMs frequently associate the artwork with positive themes such as happiness, longevity, or wealth. Consequently, both VLMs performed poorly when interpreting artworks intended to express themes related to moral integrity or societal harmony.

\subsubsection{Error Analysis}
In this section, we conduct a deeper analysis of the key observations made by the experts. Our discussion addresses the following three questions:

\textbf{(1) Is computer vision a bottleneck for understanding artworks?} We evaluated the models on text-only questions, providing only the story name conveyed by each artwork for symbolic matching. Detailed answers are listed in Appendix. Each model achieved over 80\% accuracy. However, when images were included, accuracy dropped to below 45\% for all models. These results suggest that while the models can understand the meaning of the story, they struggle to visualize what the story looks like or is composed of when interpreting the actual artwork.

\textbf{(2) What is the model's preference in understanding?} We analyzed the label-wise performance and the confusion matrix for incorrect symbolic matching answers for GPT-4o, as detailed in Appendix. GPT-4o achieves the lowest performance on options related to moral integrity and societal harmony, with accuracies around 20\%, mirroring expert observations. The confusion matrix shows that the model tends to favor option D, which relates to fecundity, among the erroneous choices.

\textbf{(3) Would fine-tuning help?} We create a custom GPT-4V to explore. We compiled 68 pieces of artwork with their annotations into a single document and uploaded it to the ChatGPT web page to build a customized Pun Rebus GPT-4V model. Since the model could only be accessed through the web page, we conducted a small-scale evaluation with 120 different artworks. The customized GPT-4V achieved a symbolic matching accuracy of 66\%, demonstrating the potential benefits of fine-tuning. The model can be accessed through \underline{\href{https://chatgpt.com/g/g-x4l7DQ2B8-pun-rebus}{this link}}, and we encourage readers to give it a try.

% \vspace{-5mm}
\section{Related Works}
% \vspace{-2mm}
\label{sec:related_works}
\subsection{Multi-modal Multicultural Understanding}
\vspace{-2mm}
Recent advancements in VLMs have spurred interest in enabling models to interpret culturally rich content. Researchers have begun to evaluate cultural commonsense~\cite{shen2024understanding}, culturally diverse facts~\cite{keleg2023dlama, Hu2023DoLL}, and cultural moral norms~\cite{ramezani2023knowledge} in LLMs. These works discover LLMs have limited culturally specific knowledge and frequently output culturally biased responses to human prompts. Some studies on multicultural visual recognition have explored improving recognition performances for food \cite{min2023large}, heritage \cite{becattini2023viscounth}, and clothing \cite{hsiao2021culture} in culturally diverse contexts. However, these works primarily focus on enhancing cultural understanding within a single modality. A more relevant effort to our proposed dataset is the MaRVL dataset \cite{liu2021visually}, aiming to evaluate multicultural reasoning abilities in VLMs. 
% Beyond incorporating multicultural reasoning in everyday contexts, previous work has not extensively explored using art as a benchmark for cultural understanding. Art is a powerful and creative medium that implicitly or explicitly reflects values, beliefs, and concerns of a culture. This makes our Pun Rebus Art dataset a unique resource for facilitating cross-cultural multimodal understanding.

\vspace{-3mm}
\subsection{Computational Pun and Pun Rebus Understanding}
\vspace{-2mm}
Computational pun understanding has been extensively studied in NLP in the last decade, with efforts made to design language models for pun detection and comprehension~\cite{zhou-etal-2020-boating, sun-etal-2022-expunations}.
More recently, researchers have investigated the abilities of LLMs in understanding puns \cite{xu2024good}, demonstrating their capability to recognize and explain puns, although generating humorous puns remains challenging. However, the understanding of pun rebus, which requires both visual recognition and language reasoning, has not been extensively studied in evaluating VLMs. To the best of our knowledge, the closest work related to our proposed dataset is the humor understanding from the image presented in \cite{hessel2023androids}, which shows that VLMs struggle to recognize the humorous elements of the visual content. 

\vspace{-2mm}
\section{Limitations}
\label{sec:limitaion}
\vspace{-2mm}

While our step-by-step error analysis provides valuable insights into the performance of VLMs on pun rebus understanding, it lacks an in-depth examination regarding the nuanced mechanisms within pun rebuses that may influence model performance. For example, we have not analyzed how the attribution of the elements (e.g., quantities, positions, etc.) in the artwork affects the models' reasoning abilities. We plan to continue collaborating with art historians to annotate each sample in the dataset with mechanism details and address this analysis in future studies.
% Although our current Chinese Pun Rebus Art dataset includes a popular and widespread convention in many Chinese decorative arts, it does not cover the complete spectrum of the rich and varied methodologies used to create traditional Chinese art, such as Chinese acrostic poem art. Therefore, our datasets only reflect a subset of the cultural expressions and artistic practices that characterize Chinese art. Therefore, the insights of VLMs derived from our study may not fully capture their strength and limitations in interpreting the full spectrum of Chinese arts.
%
Additionally, our database contains a substantial collection of ceramic arts, which are 3D objects. However, we have only used the front image for testing, thereby ignoring their 3D characteristics. Addressing this limitation is crucial for a comprehensive understanding of these artworks. We plan to incorporate the 3D aspects of these objects in our future studies.
Moreover, the expression understanding results were primarily reviewed by expert judges. While this ensures a high level of expertise, it is worth incorporating more crowdsourcing efforts to evaluate VLM's explanations to understand how different groups perceive VLM answers. This would further help identify discrepancies in understandings between experts and non-experts, shedding light on potential biases in VLM outputs.

% \begin{CJK*}{UTF8}{gbsn}
% Lastly, our designed tasks do not cover all types of punning mechanisms in Chinese art, such as those pun rebuses related to the attributions (e.g., quantities) of the visual elements. For example, the Chinese painting "three gibbons capture egrets." indicates the pun on "三猿得鹭" (sān yuán dé lù) \footnote{\url{https://www.metmuseum.org/art/collection/search/40071}} connects homophones of "three-三" (sān), "gibbons-猿" (yuán), and "egret-鹭" (lù) to represent the wish to perform well on exams, also known as "三元得路" (sān yuán dé lù), requiring the VLMs to not only recognize the objects but the also \textit{the quantities of relevant objects} in the painting to deliver the proper interpretation of the punning mechanisms and its auspicious meaning.
% \end{CJK*}

% % 
\vspace{-2mm}
\section{Conclusions}
\label{sec:conclusion}
\vspace{-2mm}

In this work, we offer the Pun Rebus Art dataset and evaluate whether state-of-the-art VLMs can interpret Chinese culture and artworks. Our findings reveal that: 
1) Current VLMs struggle to spot the salient visual elements in the Chinese Pub Rebus Arts, though they outperform ordinary humans; 
2) Due to the knowledge gap in cultural understanding, VLMs face challenges in transferring the spotted elements into their underlying auspicious meaning or matching the symbolic meanings;
3) We also observe substantial limitations in VLM's ability to provide coherent explanations for interpreting Chinese Pun Rebus Arts. The responses provided by these VLMs often exhibit biases towards fixed objects and include significant hallucinations;
4) In-context learning does not effectively guide VLMs to improve their performance in pun rebus art understanding.

In the future, a promising area of research will be developing effective data curation to incorporate more diverse and cross-cultural knowledge into the training and evaluation processes of VLMs. This approach holds promise for making VLMs more inclusive and universally beneficial, enhancing their ability to understand and interpret various cultures.

% In this work, we explore the ability of VLMs to recognize visual elements, match their symbolic meanings, and explain their understanding of Chinese Pun Rebus Arts. Specifically, we identify that:
% 1) Our results demonstrate that the current VLMs struggle to identify the relevant visual elements in the Chinese Pub Rebus Arts but are able to outperform ordinary humans; 2) Focusing on matching symbolic meanings, we observe that today's VLMs face similar challenges in recognizing the underlying auspicious meaning; 3) We also observe substantial limitations in VLM's ability to provide coherent explanations for interpreting Chinese Pun Rebus Arts. The responses provided by these VLMs often exhibit biases towards fixed objects and hallucinations; 4) We identify that in-context learning cannot effectively guide the VLMs to improve their performance in recognizing the relevant visual elements or matching the right symbolic meanings. 
% In the future, a promising area of research will be developing effective data curation to incorporate more diverse and cross-cultural knowledge into the training and evaluation processes of VLMs. This holds promise for allowing greater inclusiveness of VLMs to make these models more universally beneficial.

\bibliographystyle{plain}
\bibliography{mybib}

\newpage
\appendix
\section{Appendix}
\subsection{Datasheets for Pun Rebus Art Dataset}
\textbf{Motivation of the Dataset.}
The Pun Rebus Art dataset is designed as a comprehensive benchmark for exploring the intersection of image analysis, morphological variation, and phonological elements within the context of Chinese linguistics and cultural artifacts. This dataset is the result of extensive efforts to curate a diverse array of historical artwork documents. 

\textbf{Creator of the Dataset.}
The Pun Rebus Art Dataset was created and collected by Dr. Ni Yibin, a co-author of this paper. 

\textbf{Composition of the Dataset.}
Initiated in 1987 by Dr. Ni Yibin, a co-author of this paper, the dataset’s preparation involved meticulous collection, annotation, and verification processes that require expert knowledge of Chinese art, literature, history, and linguistics. The corpus comprises 1,011 captioned images sourced predominantly from globally-renowned Chinese-art-collecting institutions, including the Palace Museum, the Metropolitan Museum of Art, and the British Museum. Spanning over two millennia, from the Han Dynasty (206 BCE – 220 CE) to the 20th century, the dataset encompasses a rich diversity of more than ten different media types, including paintings, ceramics, bronzes, sculptures, jade, Cloisonné, lacquerware, and embroidery. The images of these artworks are stored in the dataset in the Joint Photographic Experts Group (JPEG) format.

\textbf{Distribution of the Dataset.} The Pun Rebus Art dataset could be accessed via \url{http://niyibin.org/punrebus/punrebus_main_en.php}. The code for reproducing the results of this paper is available on \url{https://github.com/zhang-tuo-pdf/Pun-Rebus-Art-Benchmark/tree/main}. It is worth noting that the category information for each data sample is stored in the GitHub link.

\textbf{Maintenance of the Dataset.} The collection of the Pun Rebus Art dataset is ongoing as we continue to curate it with additional artworks to enhance its representational diversity. We welcome researchers and enthusiasts interested in this program to join us in expanding and improving this valuable resource.

\textbf{Licence of the Dataset.} The images and their annotation in this dataset are subject to the Creative Commons Zero (CC0) license.

\subsection{Symbolic Imagery Mechanism}
In this section, we briefly describe the mechanisms behind pun rebus in visual artworks. Through our investigation, we have identified and summarized 12 distinct mechanisms that form a symbolic imagery as followings:

\textbf{Symbolic.} Using the images of people/objects in the artwork as symbols.

\textbf{Pun.} Using homophones of names of people/objects in the artwork.

\textbf{Shape.} Using the shape attributes of objects in the artwork.

\textbf{Length/Size.} Using the length or size attributes of objects in the artwork.

\textbf{Color.} Using the color attributes of objects in the artwork.

\textbf{Figure.} Using the names of people/objects in the artwork.

\textbf{Alias.} Using aliases and polyphonic characters for people/objects in the artwork.

\textbf{Numeral.} Using the quantity of visual elements in the artwork.

\textbf{Verb.} Using verbs triggered by specific actions in the artwork events.

\textbf{Preposition.} Using prepositions triggered by spatial relationships in the artwork events.

\textbf{Character.} Using pictographic Chinese characters appearing in the artwork.

\textbf{Loanword.} Using borrowed Chinese characters or radicals from the names of people/objects appearing in the artwork (names of people/objects in the artwork sound the same and share characters with the intended meaning) 

In our future work, we plan to label each sample with its corresponding mechanism and further investigate the sensitivity of VLMs to each specific mechanism.

\subsection{Experiment Details}

\textbf{LVM API Checkpoints.} For all models listed in this work, we utilize the latest model checkpoint available at the time of writing this paper. Specifically, for GPT-4o, we used \texttt{gpt-4o-2024-05-13} model; for GPT-4V, we used \texttt{gpt-4-vision-preview} model. For Gemini model, we used Gemini 1.0 Pro Vision. For Claude 3 model family, we used \texttt{claude-3-opus-20240229}, \texttt{claude-3-sonnet-20240229}, and \texttt{claude-3-haiku-20240307}. For Qwen-VL model family, we used \texttt{qwen-vl-plus} and \texttt{qwen-vl-max}.

\textbf{Computing Infrastructure.} All experiments are performed on two computing servers with ten GPUs. The server is equipped with AMD EPYC 7502 32-Core Processor and 1024G memory. The GPU is NVIDIA RTX A100. For models with API access, we just run the inference with CPUs.

\textbf{Evaluation Prompts.} For symbolic matching task, we used the following prompt for all models:
\begin{adjustwidth}{1cm}{1cm}
\begin{lstlisting}[breakatwhitespace=true, breakindent=0pt, xleftmargin=0pt]
This is a traditional Chinese artwork that likely conveys its ideas, thoughts, or wishes through symbolic, punning, shape, color, figure, numeral, verb, preposition, character, loanword or alias through the artwork. \
Carefully analyze the visual elements present in the artwork and select the option from the list below that best aligns with its conveyed meaning: \n \
    A. Longevity and Good Health \n \
    B. Happiness,Joy, Good Luck \n \
    C. Prestige, Promotion, and Good Exam Results \n \
    D. Fecundity, Harmonious Relationship and Family \n \
    E. Wealth or Prosperity \n \
    F. Moral Integrity, Eremitism \n \
    G. Peace and Protection from Evil, Societal Harmony \n \
You must make a selection using the option above in your response. Your response should start with the chosen letter that best matches the word's meaning based on a precise and sound justification for your selection. Please do not include your justification in your response.

\end{lstlisting}
\end{adjustwidth}

For element identification, we used the following prompt for all models:

\begin{adjustwidth}{1cm}{1cm}
\begin{lstlisting}[breakatwhitespace=true, breakindent=0pt, xleftmargin=0pt]
Please analyze the provided image carefully to identify key visual elements. Focus on components that traditionally have symbolic meaning in the cultural context from which the artwork originates.\
Look for elements that might represent ideas, virtues, or wishes, especially those commonly found in nature or historical motifs.\
For instance, in Chinese culture, certain animals and plants are known to symbolize specific messages when depicted in art. \
Based on these principles, identify the primary visual elements in the image that are likely used to convey a message or a wish.\
Please list the discernible elements present in the image, excluding any assumptions about elements not clearly visible.\
Pleas answer the question in one line with the following format strictly: name of element A, name of element B, etc

\end{lstlisting}
\end{adjustwidth}

For expression understanding, we used the following prompt for all models:

\begin{adjustwidth}{1cm}{1cm}
\begin{lstlisting}[breakatwhitespace=true, breakindent=0pt, xleftmargin=0pt]
This is a traditional Chinese artwork that likely conveys its ideas, thoughts, or wishes through symbolic, punning, shape, color, figure, numeral, verb, preposition, character, loanword or alias through the artwork. \
Carefully analyze the visual elements present in the artwork and select the option from the list below that best aligns with its conveyed meaning: \n \
    A. Longevity and Good Health \n \
    B. Happiness,Joy, Good Luck \n \
    C. Prestige, Promotion, and Good Exam Results \n \
    D. Fecundity, Harmonious Relationship and Family \n \
    E. Wealth or Prosperity \n \
    F. Moral Integrity, Eremitism \n \
    G. Peace and Protection from Evil, Societal Harmony \n \
You must make a selection using the option above in your response. Your response should start with the chosen letter that best matches the word's meaning, followed by a precise and sound justification for your selection.
\end{lstlisting}
\end{adjustwidth}

For text-only understanding evaluation, we used the following prompt for all models:

\begin{adjustwidth}{1cm}{1cm}
\begin{lstlisting}[breakatwhitespace=true, breakindent=0pt, xleftmargin=0pt]
f"What does the word \"{chinese_word}\" want to represent in Chinese culture? Please select the option from the list below that best aligns with its conveyed meaning: \n \
    A. Longevity and Good Health \n \
    B. Happiness,Joy, Good Luck \n \
    C. Prestige, Promotion, and Good Exam Results \n \
    D. Fecundity, Harmonious Relationship and Family \n \
    E. Wealth or Prosperity \n \
    F. Moral Integrity, Eremitism \n \
    G. Peace and Protection from Evil, Societal Harmony \n \
You must make a selection using the option above in your response. Your response should start with the chosen letter that best matches the word's meaning, followed by a precise and sound justification for your selection.
\end{lstlisting}
\end{adjustwidth}

To make sure the output answers are in a unified format for scoring, we have to made some slightly changes in words for the prompt that we used in Qwen model family. The detailed prompt that we used in experiment are listed in our GitHub link.

\begin{figure}[h]
 \centering
 \includegraphics[width = \linewidth]{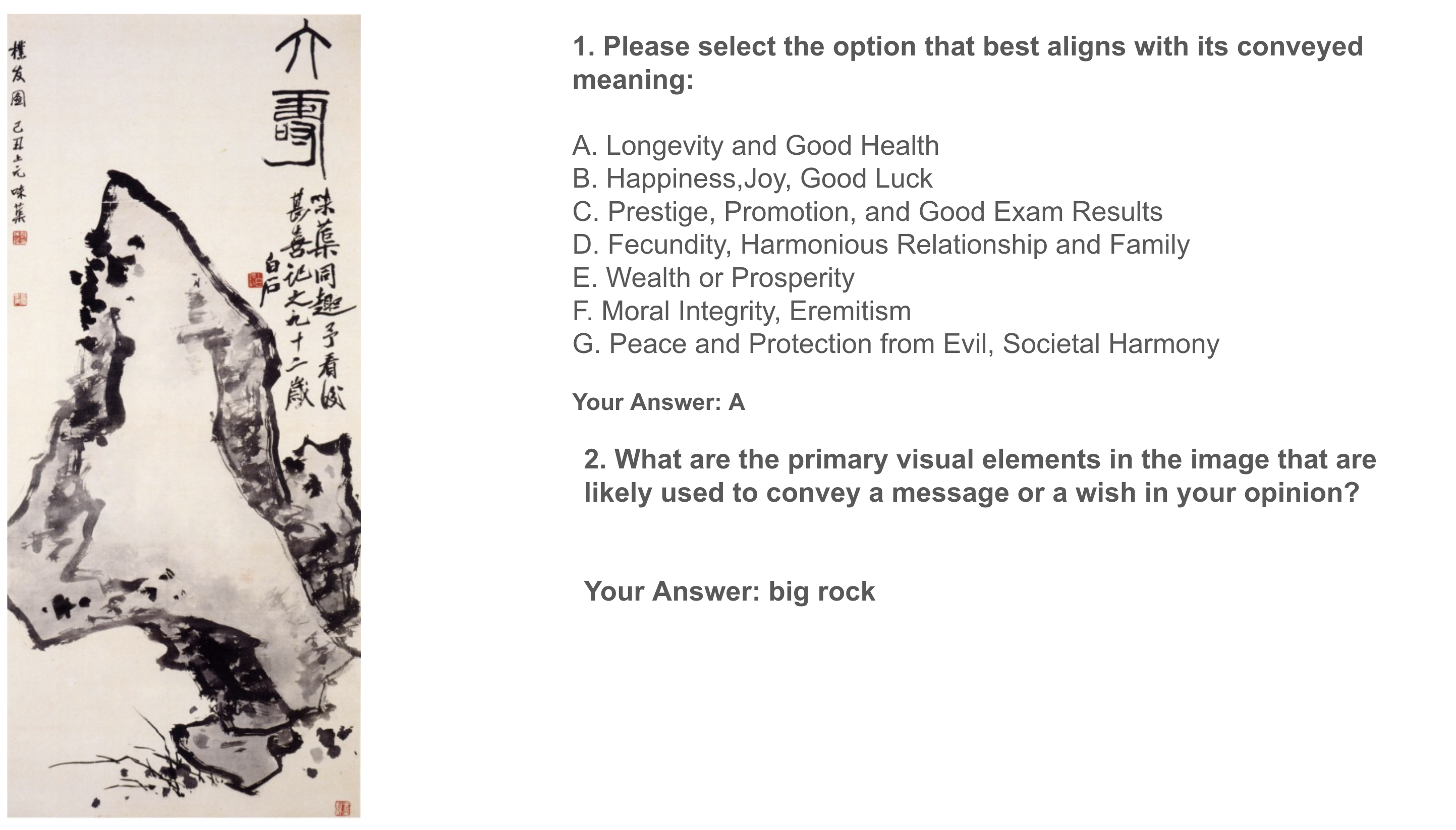}
    % \vspace{-4mm}
\caption{The example questionnaire for an artwork image to the crowd-workers. The first question is related to the symbolic matching task, and the second question is related to the element identification task.}
\label{fig:crowd}
\end{figure}

\textbf{Questions to the Crowd-workers.} 
Figure~\ref{fig:crowd} shows an example questionnaire for an artwork image to our recruited crowd-workers. We do not record any crowd-worker IDs in our experiment records. The average time of for each human evaluation is around 90 minutes, and we pay each crowd-worker \$30 each hour.
Crowdworking studies involving standard computer vision corpora (with no personal disclosures) do not require IRB review according to our institution's guidelines. Although we are not legal experts and this is not legal advice, this opinion is based on United States federal regulation 45 CFR 46, under which this study qualifies as exempt.

\subsection{Further Analysis on Experiment Results}

\textbf{Text-only Evaluation Performance.} We evaluated the models on text-only questions, providing only the story name conveyed by each artwork for symbolic matching. We used the accuracy as the evaluation metrics, the same as we used for symbolic matching task with artwork images in the main paper. The evaluation results are listed in Table~\ref{tab:text_results}.

\begin{table*}[h]
\centering
\resizebox{0.5\textwidth}{!}{
\begin{tabular}{lc}
& \multicolumn{1}{c}{Text-only Symbolic Matching} \\
\cmidrule(lr){2-2}
& Accuracy ($\uparrow$) \\
\midrule
Random Choice & 14.29\% \\
\midrule
GPT-4o & \textbf{88.55\%} \\
GPT-4V & 87.47\% \\
Gemini Pro & 85.06\% \\
Claude 3 Opus & 85.87\% \\
Claude 3 Sonnet & 85.60\% \\
Claude 3 Haiku & 86.93\% \\
Qwen-VL-Max & 84.00\% \\
Qwen-VL-Plus & 81.87\% \\
\bottomrule
\end{tabular}
}
\caption{Evaluation results for the text-only symbolic matching tasks among various VLMs. \textbf{Bold} results are best for zero-shot evaluation.
} 
\label{tab:text_results}
\end{table*}

\begin{figure}[t]
 \centering
 \includegraphics[width = \linewidth]{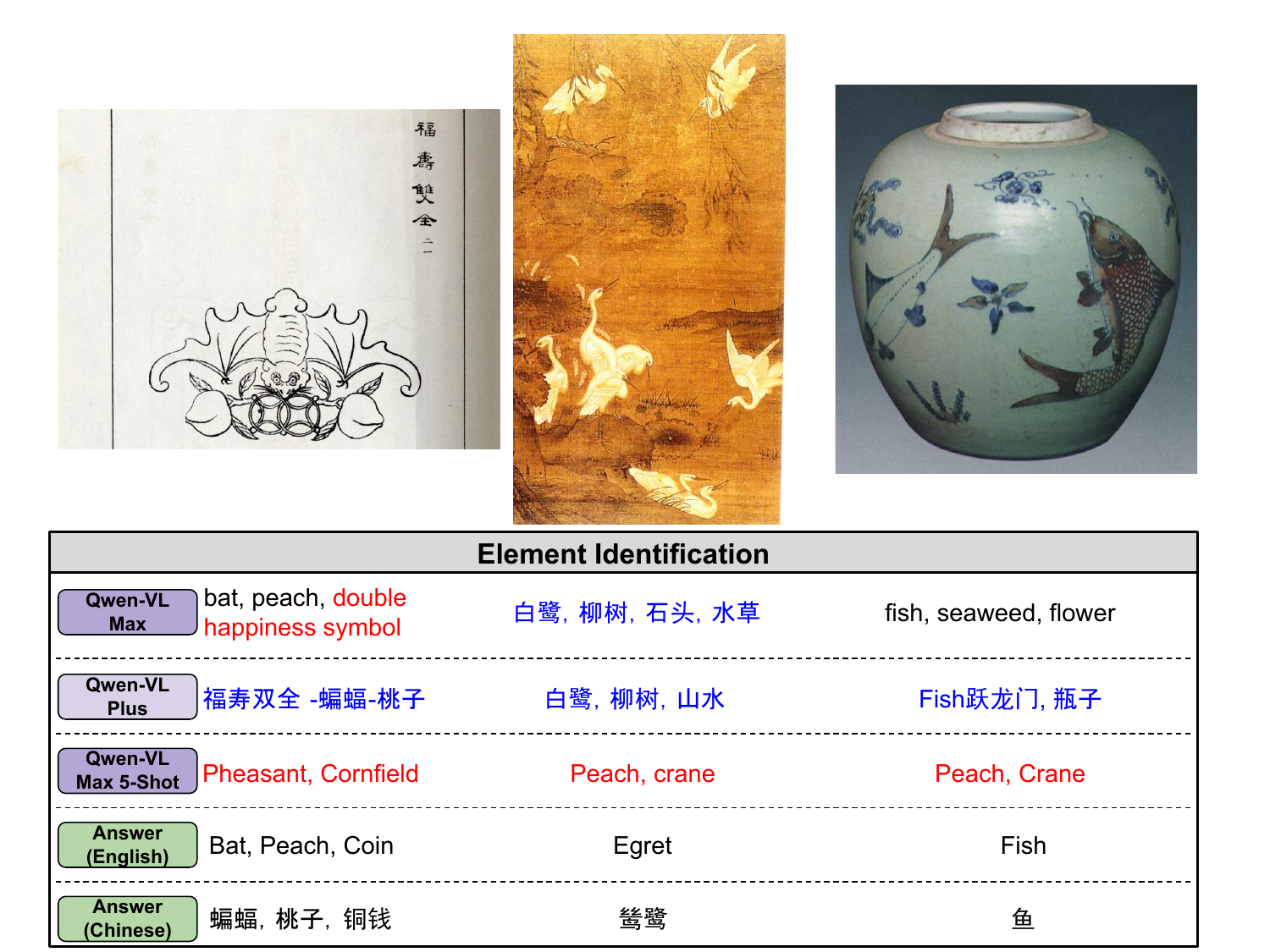}
    % \vspace{-4mm}
\caption{Several element identification examples by Qwen-VL model family. \textcolor{red}{Red} text color indicates the wrong identification results, and \textcolor{blue}{Blue} text color indicates language mismatch responses.}
\label{fig:error}
\end{figure}

\textbf{Error Examples by Qwen-VL.} As we mentioned in the Section~\ref{sec:main_results}, we observed the language mismatch in the response from Qwen-VL model family. We also observed the hallucination in the responses from Qwen-VL Max model under the 5-shot settings. In Figure~\ref{fig:error}, we provide several error examples to illustrate them. 

\begin{figure}[h]
 \centering
 \includegraphics[width = \linewidth]{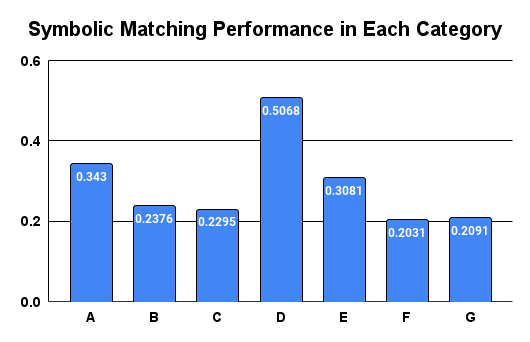}
    % \vspace{-4mm}
\caption{The category-wise accuracy performance of symbolic matching answers for GPT-4o.}
\label{fig:category}
\end{figure}

\begin{figure}[h]
 \centering
 \includegraphics[width = \linewidth]{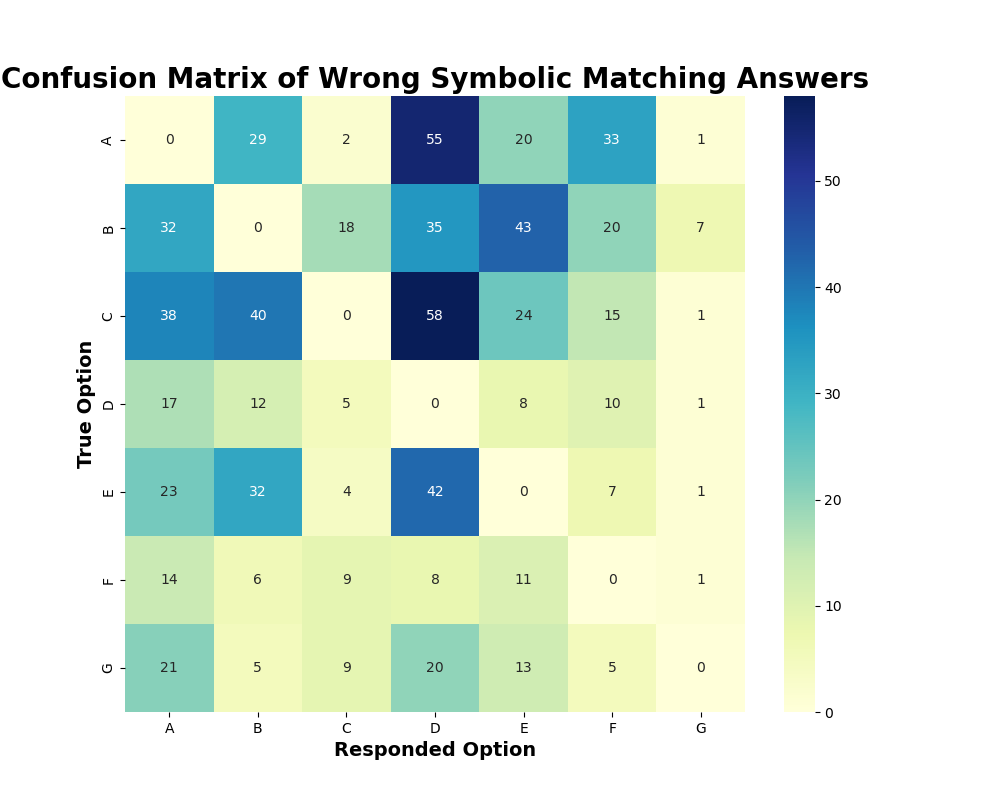}
    % \vspace{-4mm}
\caption{The confusion matrix of incorrect symbolic matching answers for GPT-4o.}
\label{fig:confusion}
\end{figure}

\textbf{Detailed Analysis on GPT-4o Results.} We analyzed the category-wise accuracy performance and the confusion matrix of incorrect symbolic matching answers for GPT-4o, as shown in Figure~\ref{fig:category} and Figure~\ref{fig:confusion}, respectively. The results indicates that GPT-4o has the most confidence on the option D, which is related to fecundity, when reading the pun rebus artwork, as it achieves the highest accuracy for this category and frequently mislabeling other answers as option D. Also, GPT-4o made its lowest accuracy on the option F and G, which are related to the moral integrity and societal harmony. The confusion matrix suggests that GPT-4o has very sparse knowledge regarding option F, as the error distribution for this category is nearly uniform compared to the errors for other options.

% \begin{figure}[h]
%   \begin{minipage}[b]{.49\textwidth}
%   \centering
%   \includegraphics[width=\textwidth]{figures/label_performance.png}
%   \caption{The category-wise accuracy performance of symbolic matching answers for GPT-4o.}
%   \label{fig:category}
%   \end{minipage}%
%   \hfill
%   \begin{minipage}[b]{.49\textwidth}
%   \centering
%   \includegraphics[width=\textwidth]{figures/heatmap_mc_gpt4o.png}
%   \caption{The confusion matrix of incorrect symbolic matching answers for GPT-4o}
%   \label{fig:confusion}
%   \end{minipage}%
% \end{figure}

\end{document}